\documentstyle[named,DINA4,twocolumn,tree-dvips]{article}

\begin{document}


%


\newlength{\texteqlength}%
\setlength{\texteqlength}{\textwidth}%
\addtolength{\texteqlength}{-3em}%

\newenvironment{tequation}%
{\begin{equation}%
\begin{minipage}[c]{\texteqlength}\begin{center}}%
{\end{center}\end{minipage}\end{equation}}%


\newenvironment{code}{\begin{tt}}{\end{tt}}

\newenvironment{fs}%
{$\left[\hspace{.5ex}\begin{tabular}{l@{\hspace{1ex}:\hspace{1ex}}l}}%
{\end{tabular}\hspace{.3ex}\right]$}%

\newlength{\cufindent}
\setlength{\cufindent}{3ex}

\newcommand{\codevspace}{\par\vspace*{0.3em}\noindent}%

\newenvironment{cuf}%
{$\left(\hspace{-0.1em}\begin{tabular}{l@{\hspace{1ex}:\hspace{1ex}}l}}%
{\end{tabular}\hspace{-0.1em}\right)$}%

\newcommand{\cufoneline}[1]{\multicolumn{2}{l}{#1}}

\newenvironment{cuflist}%
{\begin{math}\left[\begin{tabular}{l}}%
{\end{tabular}\right]\end{math}}%

\newcommand{\emptycuflist}{\begin{cuflist}$\!\!\!\!$\end{cuflist}}

\newcommand{\cuflistcons}[2]
  {\begin{math}
     \left[\begin{tabular}{c}
       \begin{math}\left.\begin{tabular}{c}
         \hspace*{-1em}#1
       \end{tabular}\right|\end{math} #2%
     \end{tabular}\right]
   \end{math}}

\newcommand{\cufsort}[2]{#1$\left(\begin{tabular}{c}#2\end{tabular}\right)$}

\newcommand{\cufclause}[2]%
{#1 := \\%
\hspace*{\cufindent}#2.}%

\newcommand{\shortcufclause}[2]{#1 := #2.}%

\newcommand{\cuftypedecl}[2]{#1 -> #2.}%

\newcommand{\cuffeatdecl}[2]
   {\begin{tabular}{l}#1 ::\\ \hspace{\cufindent}\begin{feats}#2\end{feats}\end{tabular}}%

\newcommand{\featdecl}[2]
  {\begin{tabular}{c}
     #1 \\
     \fbox{\begin{feats}#2\end{feats}}
   \end{tabular}}

\newenvironment{feats}{\begin{tabular}{l@{\hspace{1ex}:\hspace{1ex}}l}}%
{\end{tabular}}%

\newcommand{\cufenumtype}[2]{#1 = \{ #2 \}}

\newcommand{\cufsubsort}[2]{#1 < #2}

%


\newcounter{maxnode} 
\setcounter{maxnode}{500}
\newcounter{nodea}
\setcounter{nodea}{1}
\newcounter{nodeb}
\setcounter{nodeb}{2}
\newcounter{nodec}
\setcounter{nodec}{3}
\newcounter{noded}
\setcounter{noded}{4}
\newcounter{nodee}
\setcounter{nodee}{5}
\newcounter{nodef}
\setcounter{nodef}{6}
\newcounter{nodeg}
\setcounter{nodeg}{7}
\newcounter{nodeh}
\setcounter{nodeh}{8}
\newcounter{nodei}
\setcounter{nodei}{9}
\newcounter{nodej}
\setcounter{nodej}{10}
\newcounter{numberofnodes}   
\setcounter{numberofnodes}{10}  

\newcommand{\newnodenames}{
  \addtocounter{nodea}{\thenumberofnodes}
  \addtocounter{nodeb}{\thenumberofnodes}
  \addtocounter{nodec}{\thenumberofnodes}
  \addtocounter{noded}{\thenumberofnodes}
  \addtocounter{nodee}{\thenumberofnodes}
  \addtocounter{nodef}{\thenumberofnodes}
  \addtocounter{nodeg}{\thenumberofnodes}
  \addtocounter{nodeh}{\thenumberofnodes}
  \addtocounter{nodei}{\thenumberofnodes}
}

\newcommand{\oldnodenames}{
  \addtocounter{nodea}{-\thenumberofnodes}
  \addtocounter{nodeb}{-\thenumberofnodes}
  \addtocounter{nodec}{-\thenumberofnodes}
  \addtocounter{noded}{-\thenumberofnodes}
  \addtocounter{nodee}{-\thenumberofnodes}
  \addtocounter{nodef}{-\thenumberofnodes}
  \addtocounter{nodeg}{-\thenumberofnodes}
  \addtocounter{nodeh}{-\thenumberofnodes}
  \addtocounter{nodei}{-\thenumberofnodes}
                           }


%
\newcommand{\binatree}[3]{
  \newnodenames   
  \begin{array}{cc}
   \multicolumn{2}{c}{
     \node
       {\thenodea}
       {$#1$}}
   \\
   \\
    \node
      {\thenodeb}
      {$        #2
      $}
    &
   \node
      {\thenodec}
      {$
        \addtocounter{nodea}{\themaxnode}
        \addtocounter{nodeb}{\themaxnode}
        \addtocounter{nodec}{\themaxnode}
        #3
        \addtocounter{nodea}{-\themaxnode}
        \addtocounter{nodeb}{-\themaxnode}
        \addtocounter{nodec}{-\themaxnode}
      $}
  \end{array}
  \nodeconnect[b]{\thenodea}[t]{\thenodeb}
  \nodeconnect[b]{\thenodea}[t]{\thenodec}
  \oldnodenames    }

%
\newcommand{\lunatree}[2]{
      \begin{array}{c}
         #1
         \\
         |
         \\
         \mbox{{\tt #2}}
      \end{array}
   }

%
\newcommand{\unatree}[2]{
      \begin{array}{c}
         #1
         \\
         |
         \\
         #2
      \end{array}
   }


\newcommand{\verticalspace}{$\begin{array}{c}\\\end{array}\hspace*{-1em}$}

\newcommand{\vdtree}[2]{%
\newnodenames%
\begin{array}{c}
    \node
      {\thenodea}
      {$#1$}
    \\ 
    \\
    \node
      {\thenodeb}
      {\makebox[1em]{\verticalspace}}
    \node
      {\thenodec}
      {$#2$\verticalspace}
    \node
      {\thenoded}
      {\makebox[1em]{\verticalspace}}
\end{array}%
\nodeconnect[b]{\thenodea}[l]{\thenodeb}%
\nodeconnect[b]{\thenodea}[r]{\thenoded}%
\nodeconnect[l]{\thenodeb}[l]{\thenodec}%
\nodeconnect[r]{\thenodec}[r]{\thenoded}%
\oldnodenames%
}%


\newcommand{\vddottree}[2]{%
\newnodenames%
\begin{array}{c}
    \node
      {\thenodea}
      {$#1$}
    \\ 
    \\
    \node
      {\thenodeb}
      {\makebox[1em]{\verticalspace}}
    \node
      {\thenodec}
      {$#2$\verticalspace}
    \node
      {\thenoded}
      {\makebox[1em]{\verticalspace}}
\end{array}%
\makedash{3pt}%
\nodeconnect[b]{\thenodea}[l]{\thenodeb}%
\nodeconnect[b]{\thenodea}[r]{\thenoded}%
\nodeconnect[l]{\thenodeb}[l]{\thenodec}%
\nodeconnect[r]{\thenodec}[r]{\thenoded}%
\oldnodenames%
}%

\newcommand{\vddtree}[1]{%
\newnodenames%
\begin{array}{c}
    \node
      {\thenodea}
      {}
    \\ 
    \\
    \node
      {\thenodeb}
      {\makebox[1em]{\verticalspace}}
    \node
      {\thenodec}
      {$#1$\verticalspace}
    \node
      {\thenoded}
      {\makebox[1em]{\verticalspace}}%
\end{array}%
\nodeconnect[t]{\thenodea}[l]{\thenodeb}%
\nodeconnect[t]{\thenodea}[r]{\thenoded}%
\nodeconnect[l]{\thenodeb}[l]{\thenodec}%
\nodeconnect[r]{\thenodec}[r]{\thenoded}%
\oldnodenames%
}

%
\newcommand{\vdrtree}[1]{%
\newnodenames%
\begin{array}{c}
    \node
      {\thenodea}
      {$#1$}
    \\ 
    \\
    \node
      {\thenodeb}
      {\makebox[2.5em]{{}\verticalspace}}
\end{array}%
\nodeconnect[b]{\thenodea}[l]{\thenodeb}%
\nodeconnect[b]{\thenodea}[r]{\thenodeb}%
\nodeconnect[l]{\thenodeb}[r]{\thenodeb}%
\oldnodenames%
}



\newcommand{\drs}[1]{\begin{array}{|l|} \hline #1 \\ \hline \end{array}}%


\newcommand{\dref}[1]{#1}

\newcommand{\concept}[1]{\mbox{{\tt #1}}}

\newcommand{\rel}[2]{\concept{#1}(#2)}

\newcommand{\gquant}[4]{\drs{#3}\rhomb{#1}{#2}\drs{#4}}


\newcommand{\udrsdiamond}[4]{%
\begin{array}{ccc}
  & #1 &
  \\
  \\
  \node{leftq}{\begin{math}#2\end{math}} & & \node{rightq}{\begin{math}#3\end{math}}
  \\
  \\
  & \node{bottomq}{\begin{math}#4\end{math}} &
\end{array}%
\anodeconnect[t]{leftq}[b]{topscope}%
\anodeconnect[t]{rightq}[b]{topscope}%
\anodeconnect[t]{bottomq}[b]{scopeleft}%
\anodeconnect[t]{bottomq}[b]{scoperight}}%


\newcommand{\drsvspace}{\begin{tabular}{c}\mbox{}\\\mbox{}\\ \end{tabular}\hspace*{-1em}}

\newcommand{\rhomb}[2]{%
\begin{array}{ccc}%
&\node{rhombt}{}&\\%
\node{rhombl}{\drsvspace}&\begin{array}{c}#1\\#2\end{array}&\node{rhombr}{\drsvspace}\\%
&\node{rhombb}{}&%
\end{array}%
\nodeconnect[t]{rhombt}[l]{rhombl}%
\nodeconnect[t]{rhombt}[r]{rhombr}%
\nodeconnect[l]{rhombl}[t]{rhombb}%
\nodeconnect[r]{rhombr}[t]{rhombb}}%


\newcommand{\figref}[1]{Fig.\ref{#1}}

\newcommand{\natl}[1]{{\tt `#1'}}

\newcommand{\pcode}[1]{\begin{tt}#1\end{tt}}

\newlength{\codeindent} \setlength{\codeindent}{3ex}

\newcommand{\slashed}[2]{{#1}_{\mbox{\pcode{/#2}}}}

\newcommand{\dslashed}[2]{{#1}_{\mbox{\pcode{/$\!\!$/$#2$}}}}

\newcommand{\linkrel}[2] {\begin{array}{c} \node{a}{$#1$} \\ \\
\node{b}{$#2$}
 \end{array}
\makedash{1pt} \nodeconnect{a}{b} \makedash{0pt}}

%
\newtheorem{definition}{Definition}
\newtheorem{example}{Example}

\newcommand{\treepair}[2]{\left\langle\!#1#2\!\right\rangle}

\newcommand{\npair}[2]{\left\langle #1,#2 \right\rangle}

\newcommand{\stackedtrees}[2]{\!\!\begin{array}{c}#1\\[-0.8em]#2\end{array}\!\!}


\newcommand{\nullrule}[2]{
    \begin{array}{cc}
        & 
        #2
        \\[-0.7em] \cline{1-1}
        #1
    \end{array}
}

\newcommand{\unarule}[3]{
      \begin{array}{cc}
        #2
        \\[-0.7em]
        & 
        #3      
        \\[-0.7em] \cline{1-1}
        #1
      \end{array}
    }

\newcommand{\binarule}[4]{
      \begin{array}{ccc}
         #2 \;\; & \;\; #3
         \\[-0.7em]
            &    & #4 
         \\[-0.7em] \cline{1-2}
         \multicolumn{2}{c}{#1}
      \end{array}
    }



\newcommand{\lexemi}[1]{w_{#1}}


\newcommand{\grami}[1]{G_{#1}}


\newcommand{\infrule}[3]
   {#1 \;\;\stackrel{\rname{#3}}{\Longrightarrow} \;\; #2}

\newcommand{\rname}[1]{\mbox{{\it (#1)\/}}}

\newcommand{\grule}[2]
   {#1 \longrightarrow #2}

\newcommand{\sequent}[2]{#1 \rightarrow #2}

\newcommand{\genseq}[7]{\sequent{#1, \; #2, \; #3, \; #4, \; #5}{#6, \; #7}}


\newcommand{\categ}{x}
\newcommand{\categb}{y}
\newcommand{\categc}{z}
\newcommand{\categi}[1]{x_{#1}}     

\newcommand{\semi}[1]{X_{#1}}     

\newcommand{\lc}[2]{#2 \backslash #1}    
\newcommand{\rc}[2]{#1 / #2}

\title{\vspace*{-1.1cm}{\sc Syntactic-Head-Driven Generation}}
\author{\hspace*{-1em}\begin{normalsize}%
    \begin{tabular}{c}
      {\Large Esther K{\"{o}}nig}\thanks{The research reported here has been
        funded by the Sonderforschungsbereich 340 ``Sprachtheoretische
        Grundlagen f{\"{u}}r die Computerlinguistik'', a project of the
        German National Science Foundation DFG.} 
      \\ \\
      Institute for Computational Linguistics,
      Azenbergstr.12, 70174 Stuttgart,Germany, 
          esther@ims.uni-stuttgart.de
\end{tabular}%
\end{normalsize}}%
\date{}%
\maketitle%
\thispagestyle{empty}

\begin{abstract}
  The previously proposed {\em semantic\/}-head-driven generation
  methods run into problems if none of the daughter constituents in
  the syntacto-semantic rule schemata of a grammar fits the definition
  of a semantic head given in 
  \cite{Shieber/Noord/Moore/Pereira:1990}. This is the case for the
  semantic analysis rules of certain constraint-based semantic
  representations, e.g.\ Underspecified Discourse Representation
  Structures (UDRSs) 
   \cite{Frank/Reyle:1992}.

  Since head-driven generation in general has its merits, we simply
  return to a syntactic definition of `head' and demonstrate the
  feasibility of {\em syntactic\/}-head-driven generation. In addition
  to its generality, a syntactic-head-driven algorithm provides a
  basis for a logically well-defined treatment of the movement of
  (syntactic) heads, for which only ad-hoc solutions existed, so far.
\end{abstract}

\section{Introduction}
Head-driven generation methods combine both, top-down search and
bottom-up combination, in an ideal way.  
\cite{Shieber/Noord/Moore/Pereira:1990} proposed to define the `head'
constituent~$h$ of phrase with category~$x$ on {\em semantic\/}
grounds: the semantic representations of $h$ and $x$ are identical.
This puts a strong restriction on the shape of semantic analysis
rules: one of the leaves must share its semantic form with the root
node. However, there are composition rules for semantic
representations which violate this restriction, e.g.\ the schemata for
the construction of Underspecified Discourse Representation Structures
(UDRSs) 
\cite{Frank/Reyle:1992} where, in general,
the root of a tree is associated with a strictly larger semantic
structure than any of the leaves. In order to make a generation method
available for grammars which do not follow the strict notion of a
semantic head, a {\em syntactic\/}-head-driven generation algorithm is
presented, which can be specialized to generate from UDRSs. In a
second step, the method will be extended in order to handle the
movement of (syntactic) heads in a logically well-defined manner.

The (tactical) generation problem is the task to generate a string
from a semantic representation according to the syntax-semantics-relation
defined in a given grammar. Let's assume that the latter relation is
stated by pairs of trees. The left tree states a local syntactic
dependency, i.e.\ the dominance relation between a root node and a set
of leaf nodes and the linear precedence relation among the leaves. The
right tree defines the relation among the semantic representation of
the root and the semantic representations of the leaves. We assume
that there is a one-to-one map from the nonterminal leaf
nodes of the (local) syntax tree on the leaf
nodes of the (local) semantic derivation tree. Example:
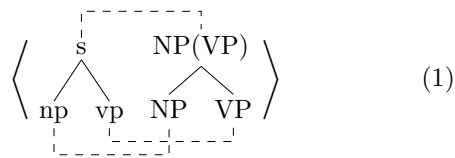
\begin{equation}
\begin{array}{c}
   \\
   \treepair
     {
  \begin{array}{cc}
   \multicolumn{2}{c}{
     \node
       {a}
       {s}}
   \\
   \\
    \node
      {b}
      {np}
    &
   \node
      {c}
      {vp}
  \end{array}
     }
     {
  \begin{array}{cc}
   \multicolumn{2}{c}{
     \node
       {d}
       {NP(VP)}}
   \\
   \\
    \node
      {e}
      {NP}
    &
   \node
      {f}
      {VP}
  \end{array}
     }
    \\ \mbox{}
\end{array}
  \nodeconnect[b]{a}[t]{b}
  \nodeconnect[b]{a}[t]{c}
  \nodeconnect[b]{d}[t]{e}
  \nodeconnect[b]{d}[t]{f}
  \makedash{3pt}
  \barnodeconnect[1em]{a}{d}
  \barnodeconnect[-1em]{b}{e}
  \barnodeconnect[-0.5em]{c}{f}
\end{equation}
If one assumes a pairwise linking from left to right then the links
between the two trees can be omitted.
Although such pairs of trees are reminiscent of synchronous trees in
TAG's 
\cite{Shieber/Schabes:1991}, they are
simpler in various ways, in particular because we will {\em not\/}
make use of the adjunction operation later on. In essence, pairs of
trees are just a graphical notation for what has been put forward as
the `rule-to-rule'-hypothesis, cf.\ 
\cite{Gazdar/Klein/Pullum/Sag:1985}, the fact that in the grammar
each syntax rule is related with a semantic analysis rule. However, on
the long run, the tree notation suggests a more general relation,
e.g.\ more internal structure or additional, terminal leaf nodes in
the local syntax tree.

\begin{figure*}
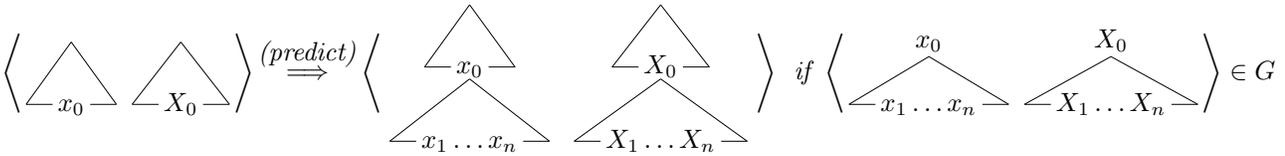

all leaves of the syntax tree are labeled with terminals
\hspace{1cm}
\rname{success} 
\\[1em]
\begin{math}
\infrule
  {\treepair
     {\vddtree
        {\categi{0}}}
     {\vddtree
        {\semi{0}}}\hspace*{-0.7em}}
  {\hspace*{-0.7em}\treepair
     {\stackedtrees
        {\vddtree
           {\categi{0}}}
        {\vddtree
           {\categi{1} \ldots \categi{n}}}}
     {\stackedtrees
         {\vddtree
            {\semi{0}}}
         {\vddtree
            {\semi{1} \ldots \semi{n}}}}}
   {predict}
 \begin{array}{l}
   \mbox{{\em if\ }} 
   \treepair
     {\vdtree
        {\categi{0}}
        {\categi{1} \ldots \categi{n}}}
     {\vdtree
        {\semi{0}}
        {\semi{1} \ldots \semi{n}}}
   \in \grami{}
 \end{array}
\end{math}
\caption{Top-Down Generation 
 ($\grami{}$ grammar description; 
  $\categi{i}$ syntactic category; 
  $\semi{i}$ semantic  representation)}
\label{tdgen}
\end{figure*}

An obvious way to implement a generation procedure (see
\figref{tdgen}) is to relate the input semantics with the start
symbol of the grammar and then to try to expand this node in a
top-down manner according to the rules specified in the grammar. This
node expansion corresponds to an application of the
\rname{predict}-rule in the following abstract specification of a
top-down generator. Generation terminates successfully if all the leaf
nodes are labeled with terminals \rname{success}.  The question is
which method is used to make two, possibly complex symbols equal. For
the sake of simplicity, we assume that the open leaves $\categi{0}$
resp.\ $\semi{0}$ are matched by (feature) term unification with the
corresponding mother nodes in the grammar rule. However, for the
semantic form~$\semi{0}$, a decidable variant of higher order
unification might be used instead, in order to include the reduction
of $\lambda$-expressions. Of course, the necessary precautions have to
be taken in order to avoid the confusion between object- and
meta-level variables, cf.\ \cite{Shieber/Noord/Moore/Pereira:1990}.

A depth-first realization of this abstract top-down algorithm would
work fine as long as the semantic representations of the leaves are
always strictly smaller in size as the semantic form of the root node.
But, if the actual semantic decomposition takes place in the lexicon,
the semantic representations of some subgoals will be variables,
which stand for semantic representations of any size:
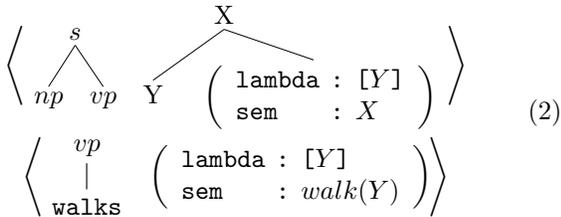
\begin{equation} \label{srule}
\begin{array}{c}
   \treepair
     {\binatree
       {s}
       {np}
       {vp}
     }
     {\begin{array}{cc}
        \multicolumn{2}{c}{\node{a}{X} \hspace{5em}}
        \\ \\
        \node{b}{Y}
        &
        \node{c}{\mbox{\begin{code}
         \begin{cuf}
           lambda & [$Y$] \\
           sem & $X$ 
         \end{cuf}\end{code}}}
      \end{array}%
\nodeconnect{a}{b}%
\nodeconnect{a}{c}}%
\\[0.5em]
   \treepair
     {\unatree
       {vp}
       {\mbox{{\tt walks}}}
     }
     {\mbox{\begin{code}
      \begin{cuf}
        lambda & 
          [$Y$] \\
        sem & $walk(Y)$ \\
      \end{cuf}\end{code}}
    }
\end{array}
\end{equation}
A strict left-to-right, depth-first expansion of subgoals might run
into problems with the grammar fragment in~(\ref{srule}) if a
left-recursive $np$-rule exists, because the semantics of the $np$ is
only instantiated once the 'semantic head' of the $vp$ has been looked
up in the lexicon.

\begin{figure*}
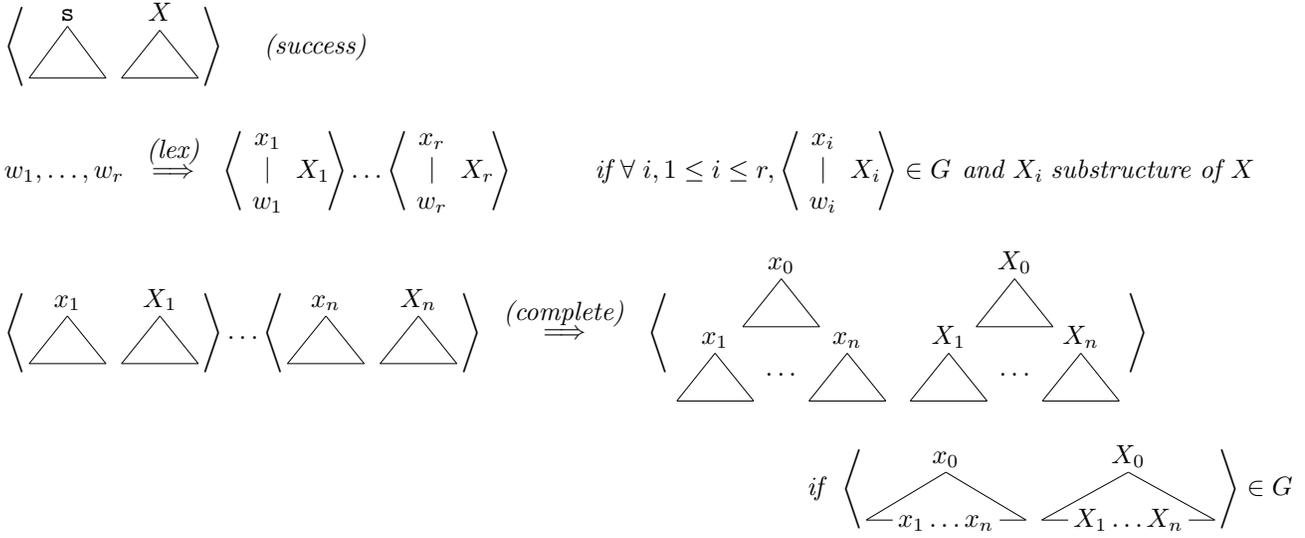

\begin{math}
  \treepair
     {
      \vdrtree
        {\pcode{s}}
     }
     {
      \vdrtree
        {\semi{}}
     }
 \;\;\;\;\;
 \rname{success} 
\end{math}
\\[1em]
\begin{math}
\infrule
  {\lexemi{1},\ldots,\lexemi{r}}
  {\treepair
     {\unatree
        {\categi{1}}
        {\lexemi{1}}
     }
     {\semi{1}}
   \ldots
   \treepair
     {\unatree
        {\categi{r}}
        {\lexemi{r}}
     }
     {\semi{r}}
   }
   {lex} 
   \hspace{3em}
   \mbox{{\em if\ }} 
   \forall \; i, 1 \leq i \leq r,
     \treepair
       {\unatree
          {\categi{i}}
          {\lexemi{i}}
       }
       {\semi{i}}
      \in \grami{} 
   \mbox{{\em \ and\ }} 
    \semi{i}\; \mbox{\em substructure of\ } \semi{}
\end{math}
\\[1em]
\begin{math}
\infrule
  {
   \treepair
     {\vdrtree
        {\categi{1}}
     }
     {\vdrtree
        {\semi{1}}
     }
   \ldots
   \treepair
     {\vdrtree
        {\categi{n}}
     }
     {\vdrtree
        {\semi{n}}
     }
  }
  {\treepair
     {
      \stackedtrees
        {\vdrtree
           {\categi{0}}
        }
        {\vdrtree
           {\categi{1}}
         \ldots
         \vdrtree
           {\categi{n}}
        }
     }
     {\stackedtrees
        {\vdrtree
           {\semi{0}}
        }
        {\vdrtree
           {\semi{1}}
         \ldots
         \vdrtree
           {\semi{n}}
        }
     }
   }
   {complete}
\end{math}
\begin{flushright}
   {\em if\ }
   $\treepair
     {\vdtree
        {\categi{0}}
        {\categi{1} \ldots \categi{n}}}
     {\vdtree
        {\semi{0}}
        {\semi{1} \ldots \semi{n}}}
   \in \grami{}$  
\end{flushright}
\caption{Bottom-Up Generation 
  ($\grami{}$ grammar description; 
   {\tt s} start symbol; 
  $\semi{}$ input semantics;  
   $\categi{i}$ syntactic category; 
   $\semi{i}$ semantic representation)}
\label{bg}
\end{figure*}

\section{Previous work}
A top-down, semantic-structure-driven generation algorithm has been
defined by 
\cite{Wedekind:1988}
which gives a basis for dynamic subgoal-reordering guided by the
semantic input. Some proposals have been made for subgoal reordering
at compile-time, e.g.\ 
\cite{Minnen/Gerdemann/Hinrichs:1993} elaborating on the work by
\cite{Strzalkowski:1990}.
But there will be no helpful subgoal reordering for rules with
semantic head recursion:
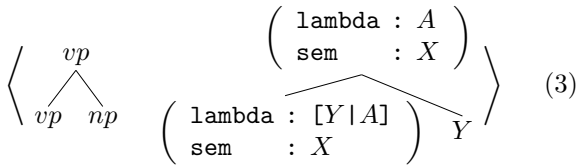
\begin{equation} 
   \treepair
     {\binatree
       {vp}
       {vp}
       {np}
     }
     {\begin{array}{cc}
        \multicolumn{2}{c}{\hspace*{4em}\node{a}{\begin{code}
         \begin{cuf}
           lambda & $A$ \\
           sem & $X$ 
         \end{cuf}\end{code}}}
        \\ \\
        \node{b}{\begin{code}
         \begin{cuf}
           lambda & [$Y$|$A$] \\
           sem & $X$ 
         \end{cuf}\end{code}}
        &
        \node{c}{$Y$}
      \end{array}%
     }
\end{equation}%
\nodeconnect{a}{b}%
\nodeconnect{a}{c}%
Obviously, a bottom-up component is required. One solution is to keep
to a top-down strategy but to do a breadth-first search, cf.\ 
\cite{Kohl:1992}, which will be
fair and not delay the access to the lexicon forever, as a pure
depth-first strategy does. 
Alternatively, one could adopt a pure bottom-up strategy like the one
which has been proposed in 
\cite{Shieber:1988} and which
is presented in \figref{bg} in a highly schematic manner.
A lexical entry qualifies as a potential leaf node if its semantic
form is a non-trivial substructure of the input semantics (rule
\rname{lex}). The derivation trees are built up by the
\rname{complete}-rule. Generation finally succeeds if the root node of
the current syntax tree is labeled with the start symbol of the grammar
and the root of the semantic analysis tree with the input semantics.
Due to the exclusion of phrases with 'empty' semantics (which would be
trivial substructures of the input semantics), the method always
terminates. However, the lack of top-down guidance will lead, in
general, to a lot of non-determinism. The strong substructure
condition means that the algorithm will be incomplete for grammars
which cover semantically void phrases like expletive
expressions, particles, and subphrases of idioms.

\begin{figure*}
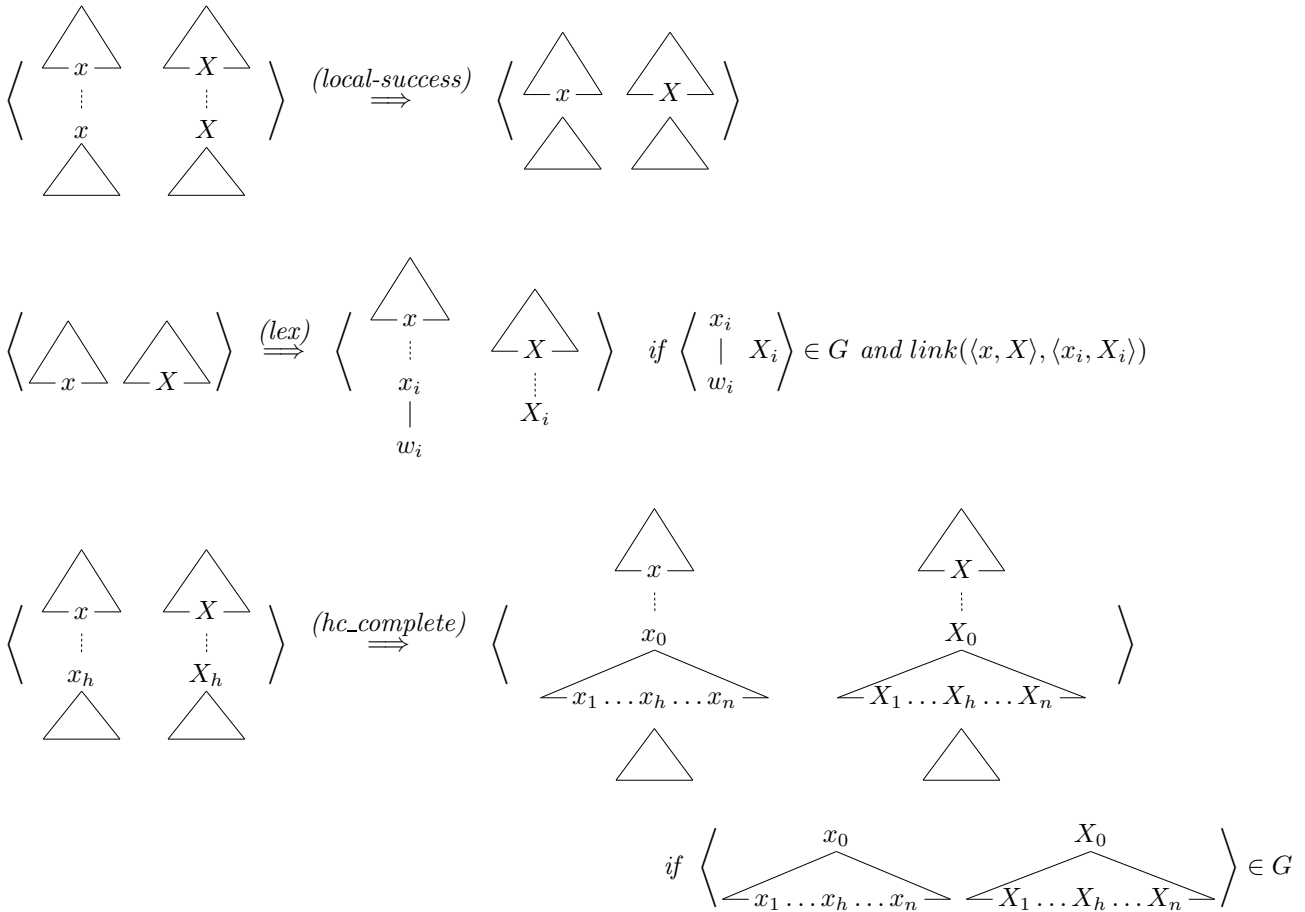

all leaves are labeled with terminals and the tree does not contain
any dotted lines
\hspace{2em}
\rname{global-success} 
\\[1em]
\begin{math}
  \infrule
    {\treepair
       {\linkrel
          {\vddtree
             {\categ}
          }
          {\vdrtree
             {\categ}
          }
       }
       {\linkrel
          {\vddtree
             {\semi{}}
          }
          {\vdrtree
             {\semi{}}
          }
       }
    }
    {\treepair
        {\stackedtrees
           {\vddtree
              {\categ}}
           {\vdrtree
              {}}
       }
       {\stackedtrees
          {\vddtree
             {\semi{}}}
          {\vdrtree
             {}}
       }
    }
    {local-success} 
\end{math}
\\[1em]
\begin{math}
\infrule
  {
   \treepair
     {
      \vddtree
        {\categ}
     }
     {
      \vddtree
        {\semi{}}
     }
  }
  {
   \treepair
     {\linkrel
        {\vddtree
           {\categ}
        }
        {
         \unatree
           {\categi{i}}
           {\lexemi{i}}
        }
     }
     {\linkrel
        {\vddtree
           {\semi{}}
        }
        {\semi{i}
        }
     }
   }
   {lex} 
   \hspace{1em}
   \mbox{{\em if\ }}
   \treepair
     {\unatree
        {\categi{i}}
        {\lexemi{i}}
     }
     {\semi{i}}
    \in \grami{} 
    \mbox{\ {\em and}\ } 
    link(\langle\categ,\semi{}\rangle,\langle\categi{i},\semi{i}\rangle)
\end{math}
\\[1em]
\begin{math}
\infrule
  {
   \treepair
     {\linkrel
        {\vddtree{\categ}}
        {\vdrtree
           {\categi{h}}
        }
     }
     {\linkrel
        {\vddtree{\semi{}}}
        {\vdrtree
           {\semi{h}}
        }
     }
  }
  {
   \treepair
     {\linkrel
        {\vddtree{\categ}}
        {
         \begin{array}{c}
           \vdtree
             {\categi{0}}
             {\categi{1} \ldots \categi{h} \ldots \categi{n}} \\[-0.5em]
           \vdrtree
             {}
         \end{array}
        }
     }
     {\linkrel
        {\vddtree{\semi{}}}
        {\begin{array}{c}
           \vdtree
             {\semi{0}}
             {\semi{1} \ldots \semi{h} \ldots \semi{n}} \\[-0.5em]
           \vdrtree
             {}
         \end{array}
       }
     }
  }
  {hc\_complete}
\end{math}
\begin{flushright}
  {\em if\ }
   $\treepair
     {\vdtree
        {\categi{0}}
        {\categi{1} \ldots \categi{h} \ldots \categi{n}}}
     {\vdtree
        {\semi{0}}
        {\semi{1} \ldots \semi{h} \ldots \semi{n}}}
   \in \grami{}$
\end{flushright}
\caption{Head-Corner Generator
  ($\grami{}$ grammar description; 
   $\categi{i}$ syntactic category; 
   $\semi{i}$ semantic representation)}
\label{hcgen}
\end{figure*}

\begin{figure*}
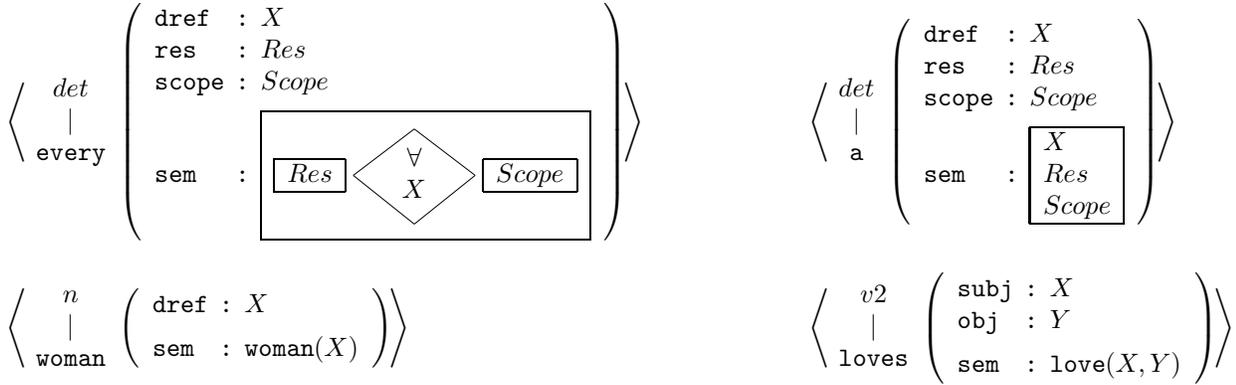

\begin{center}
\begin{code}
\begin{math}
\begin{array}{ll}
\treepair
  {\lunatree
     {det}
     {every}
  }
  {\mbox{\begin{cuf}
     dref & $X$ \\
     res & $Res$ \\
     scope  & $Scope$ \\
     \cufoneline{} \\[-0.7em]
     sem & 
      $\drs
         {\gquant
            {\forall}
            {X}
            {Res}
            {Scope}
         }$
   \end{cuf}}
  }
  \hspace*{5em}
  &
\treepair
  {\lunatree
     {det}
     {a}
  }
  {\mbox{\begin{cuf}
      dref & $X$ \\
      res & $Res$ \\
      scope  & $Scope$ \\
      \cufoneline{} \\[-0.7em]
     sem & 
      $\drs
         {\dref{X} \\
          Res \\
          Scope
         }$
   \end{cuf}}
  }
\\ \\
\treepair
  {\lunatree
     {n}
     {woman}
  }
  {\mbox{\begin{cuf}
       dref & $X$ \\
       \cufoneline{} \\[-0.7em]
       sem & 
          $\rel
             {woman}
             {X}$
      \end{cuf}}
   }
  &
\treepair
  {\lunatree
     {v2}
     {loves}
  }
  {\mbox{\begin{cuf}
      subj & $X$ \\
      obj & $Y$ \\
      \cufoneline{} \\[-0.7em]
      sem & $\rel{love}{X,Y}$
    \end{cuf}}
   }
\end{array}
\end{math}
\end{code}
\end{center}
\caption{A grammar with UDRS-construction rules - lexicon}
\label{hpsgl}
\end{figure*}

The head-corner generator in \cite{Noord:1993} is an illustrative
instance of a sophisticated combination of top-down prediction and
bottom-up structure building, see \figref{hcgen}. The rule
\rname{lex} restricts the selection of lexical entries to those which
can be `linked' to the local goal category (visualized by a dotted
line). According to van Noord, two syntax-semantics pairs are linkable
if their semantic forms are identical, i.e.\
\begin{math}
  link(\langle\categ,\semi{}\rangle,\langle\categi{i},\semi{}\rangle)
\end{math}.
The rule \rname{hc-complete} performs a 'head-corner' completion step
for a (linked) phrase~$x_{h}$, which leads to the prediction of the
head's sisters. A link marking can be removed if the linked categories
resp.\ the linked semantic forms are identical (rule
\rname{local-success}).  Generation succeeds if all the leaves of the
syntax tree are labeled with terminals and if no link markings exist
(rule \rname{global-success}).  In order to obtain completeness in the
general case, the inference schemata of the head-corner generator must
be executed by a breadth-first interpreter, since a depth-first
interpreter will loop if the semantic analysis rules admit that
subtrees are associated with semantic forms which are not proper
substructures of the input semantics, and if these subtrees can be
composed recursively.  Such an extreme case would be a recursive rule
for semantically empty particles: ('empty' semantics is represented by
the empty list symbol \pcode{[]}):
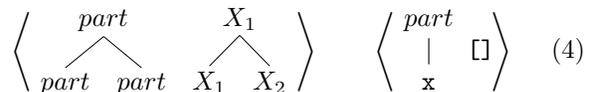
\begin{equation} 
  \treepair
     {\binatree
       {part}
       {part}
       {part}
     }
     {\binatree
        {\semi{1}}
        {\semi{1}}
        {\semi{2}}
     }
   \hspace{2em}
  \treepair
     {\unatree
       {part}
       {\pcode{x}}
     }
     {\mbox{\pcode{[]}}}
\end{equation}
However, if we assume that structures of that kind do not occur, a
depth-first interpreter will be sufficient, e.g.\ the inference rules
of the algorithm can be encoded and interpreted directly in Prolog.
Note that van Noord's method is restricted to grammars where phrases
have always a lexical semantic head. The algorithm in 
\cite{Shieber/Noord/Moore/Pereira:1990} relaxes this condition.


%
\section{Underspecified Discourse Representation Structure}
\begin{figure*}
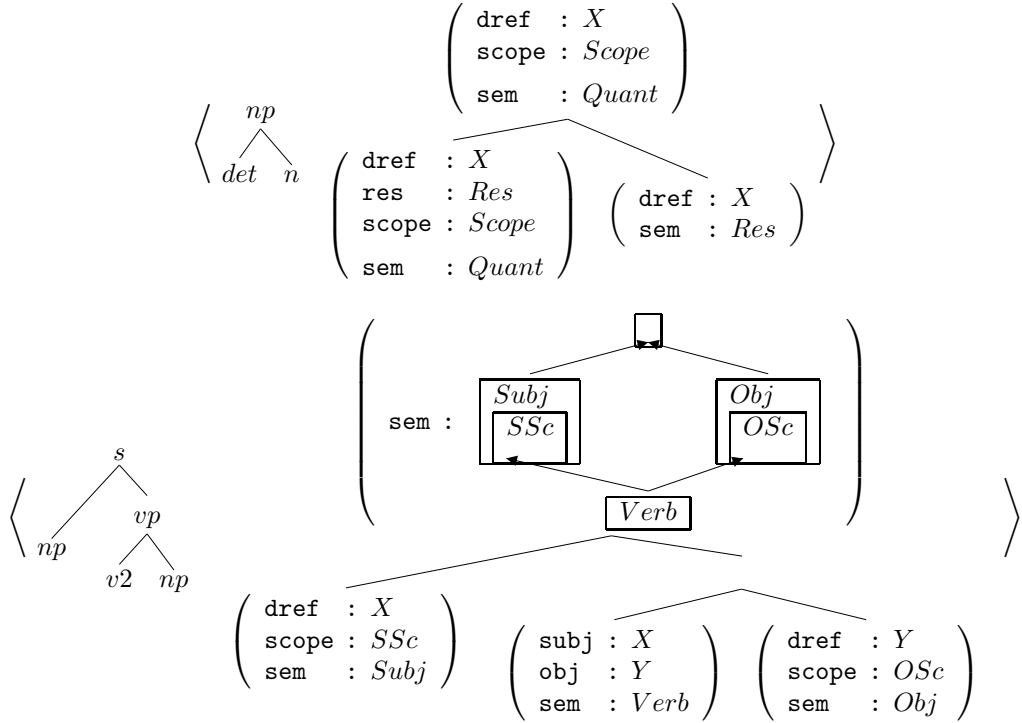

\begin{center}
\begin{code}
\begin{math}
\treepair
  {\binatree
     {np}
     {det}
     {n}
  }
  {\binatree
     {\mbox{\begin{cuf}
        dref & $X$ \\
        scope  & $Scope$ \\ 
        \cufoneline{} \\[-0.7em]
        sem & $Quant$
       \end{cuf}}
     }
     {\mbox{\begin{cuf}
        dref & $X$ \\
        res & $Res$ \\
        scope & $Scope$ \\ 
        \cufoneline{} \\[-0.7em]
        sem & $Quant$
      \end{cuf}}
     }
     {
      \mbox{\begin{cuf}
        dref & $X$ \\
        sem & $Res$
      \end{cuf}}
    }
  }
\end{math}
\end{code}
\\[1em]
\begin{code}
\begin{math}
\treepair
   {\binatree
     {s}
     {np}
     {\binatree
        {vp}
        {v2}
        {np}
     }
  }
  {
    \binatree
       {\mbox{\begin{cuf}
          sem &
           \begin{math}
             \udrsdiamond
               {\drs{\node{topscope}{\mbox{}}}}
               {\drs
                  {Subj \\
                   \drs{SSc \\[-0.5em]
                        \node{scopeleft}{\mbox{}}}}}
               {\drs
                  {Obj \\
                   \drs{OSc \\[-0.5em]
                        \node{scoperight}{\mbox{}}}}}
               {\drs{Verb}}
           \end{math}
      \end{cuf}}
      }
      {
        \mbox{\begin{cuf}
          dref & $X$ \\
          scope  & $SSc$ \\
          sem & $Subj$
        \end{cuf}}
      }
      {
  \binatree
     {}
     {
       \mbox{\begin{cuf}
         subj & $X$ \\
         obj & $Y$ \\
         sem & $Verb$        
      \end{cuf}}
     }
     {
       \mbox{\begin{cuf}
        dref & $Y$ \\
        scope  & $OSc$ \\
        sem & $Obj$
      \end{cuf}}
     }
   }
    }
\end{math}
\end{code}
\end{center}
\caption{A grammar with UDRS-construction rules - syntax rules}
\label{hpsgr}
\end{figure*}

In the following, we will present shortly a semantic representation
formalism and a corresponding set of analysis rules which resist to
the definition of `semantic head' as it is required in van Noord's
head-corner algorithm.  
\cite{Reyle:1993daucrd} developed an
inference system for {\em Underspecified Discourse Representation
  Structures\/} (UDRS's), i.e.\ Discourse Representation Structures
\cite{Kamp/Reyle:1993} which are underspecified with respect to
scope.  The following UDRS represents simultaneously the two readings
of the sentence \natl{every woman loves a man} by leaving the exact
structural embedding of the quantified phrases underspecified.
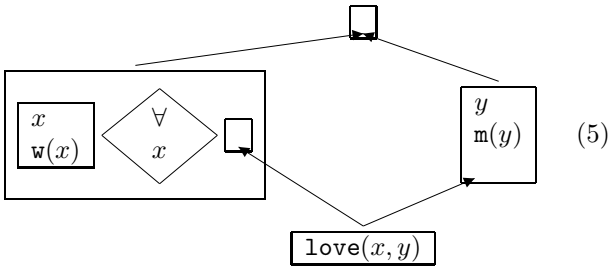
\begin{equation}
\hspace{-1em}
  \udrsdiamond
    {\drs{\node{topscope}{\mbox{}}}
    }
    {\drs
       {\gquant
          {\forall}
          {x}
          {\dref{x} \\
           \rel{w}{x}
          }
          {\node{scopeleft}{\mbox{}}
          }
       }
     }
     {\drs
        {\dref{y} \\
         \rel{m}{y} \\
         \node{scoperight}{\mbox{}}}
     }
     {\drs
        {\rel{love}{x,y}}
     }
\end{equation}
An arrow pointing from $\semi{2}$ to $\semi{1}$ is called a {\em
  subordination constraint\/} and means that the formula~$\semi{2}$
must not have wider scope than $\semi{1}$.  
\cite{Frank/Reyle:1992} proposed rules for the construction of UDRS's
in an HPSG-style syntax, cf.\ 
\cite{Pollard/Sag:1993}, which are shown in \figref{hpsgl} and~\ref{hpsgr} in a
somewhat adapted manner. Semantic composition is performed by
the coindexing of the features \pcode{dref}, \pcode{res}, \pcode{subj}, etc. 
which serve as an interface to the value of the \pcode{sem} feature,
the actual semantic representation.
For the phrase-structure tree rooted with~$s$, there is no leaf which
would fulfill the definition of a semantic head given in
\cite{Shieber/Noord/Moore/Pereira:1990} or \cite{Noord:1993}. Hence,
the head-corner generator of \figref{hcgen} with a link
relation based on semantic heads will not be applicable.

\section{Syntactic-head-driven generation}
\subsection{A new link relation}
One could define a {\em weak notion of a semantic head\/} which requires that
the semantic form of the semantic head is a (possibly empty)
substructure of the root semantics. But this is rather meaningless,
since now every leaf will qualify as a semantic head. As a way out,
there is still the notion of a syntactic head, which can serve as the
pivot of the generation process.

Assume that the syntactic head leaf for each local syntax tree has
been defined by the grammar writer.  We get the following {\em
  preliminary version of a syntax-based link relation\/}:
\begin{equation}
  link(\langle\categ,\semi{}\rangle,\langle\categi{i},\semi{i}\rangle)
\end{equation}
\begin{enumerate}
  \item
    {\em if either\/} $\categ$ = $\categi{i}$
  \item
    {\em or} $\categi{j}$ is a possible syntactic head of $\categ$

    \hspace*{1em}and 
       $link(\langle\categi{j},\semi{j}\rangle,\langle\categi{i},\semi{i}\rangle)$
\end{enumerate}
This is the kind of link relation which is used for parsing. In
general, it works fine there, because with each lexical lookup a part
of the input structure, i.e.\ of the input string, is consumed. In
order to reduce the number of non-terminating cases for generation, a
similar precaution has to be added, i.e.\ the input structure has to
be taken into account. The {\em final version of a syntax-based link
  relation\/} incorporates a test for the weak notion of a semantic
head:
\begin{equation}
  link(\langle\categ,\semi{}\rangle,\langle\categi{i},\semi{i}\rangle)
  \mbox{\ \ {\em if\/}}
\end{equation}
\begin{enumerate}
  \item
    {\em either\/} $\categ$ = $\categi{i}$ and
    
    $\semi{i}$ is a (possibly empty) substructure of $\semi{}$
  \item
    {\em or} 
       $\categi{j}$ is a possible syntactic head of $\categ$

     and 
       $link(\langle\categi{j},\semi{}\rangle,\langle\categi{i},\semi{i}\rangle)$
\end{enumerate}
The substructure check makes only sense if the semantics~$\semi{}$ of
the current goal is instantiated. This might not be the case, when
the proper semantic head and the syntactic head differ, and a sister
goal of the semantic head is to be expanded before the head itself.
Hence, in general, the sister goals must be reordered according to the
degree of instantiation of their semantic representations.  In
addition to the improved termination properties, the condition on the
semantic representation helps to filter out useless candidates from
the lexicon, i.e.\ lexical entries which will never become part of the
final derivation because their semantic representations do not fit.

\begin{figure*}
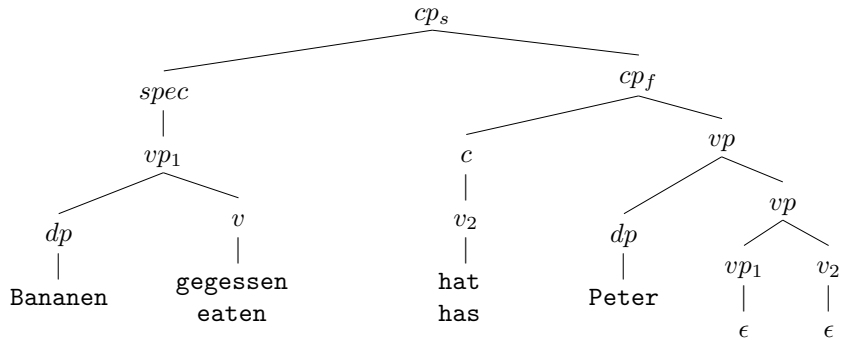

\centering
\begin{math}
\binatree
  {cp_{s}}
  {\unatree
     {spec
     }
     {\binatree
        {vp_{1}}
        {\lunatree
           {dp}
           {Bananen}
        }
        {\lunatree
           {v}
           {\begin{tabular}{c}
              gegessen \\
              eaten
           \end{tabular}
          }
        }
     }
  }
  {\binatree
     {cp_{f}}
     {
      \unatree
        {c}
        {\lunatree
           {v_{2}}
           {\begin{tabular}{c}
              hat \\
              has
           \end{tabular}
          }
       }
     }
     {\binatree
        {vp}
        {\lunatree
           {dp}
           {Peter}
        }
        {\binatree
           {vp}
           {\unatree
              {vp_{1}}
              {\epsilon}
           }
           {\unatree
              {v_{2}}
              {\epsilon}
           }
       }
     }
  }
\end{math}
\caption{Movement of a complex non-head}
\label{vtwo}
\end{figure*}

\subsection{Grammars with head movement}
In order to simplify the representation in the following, we assume
that each syntax tree in a grammar is isomorphic to the corresponding
semantic analysis tree. This means that both trees can merged into one
tree by labeling the nodes with syntax-semantics-pairs:
\begin{equation}
\binatree
  {\npair{\categi{0}}{\semi{0}}}
  {\npair{\categi{1}}{\semi{1}}}
  {\npair{\categi{2}}{\semi{2}}}
\end{equation}

In \cite{Shieber/Noord/Moore/Pereira:1990} an ad-hoc solution was
proposed to enforce termination when the semantic head has been moved.
By adopting a syntactic-head-driven strategy, head-movement does not
cause a problem if the landing site of the head is the `syntactic
head' (or rather: the main functor category of the clause, in
categorial grammar terminology) of a superordinate clause.
This is postulated by syntactic descriptions like
\begin{equation}  \label{verbf}
\hspace*{-1em}
  \binatree 
    {\npair
       {cp_{f}}
       {\semi{0}}}
    {\unatree 
       {c} 
       {\npair{v_{i}}{\semi{1}}}
    } 
    {\slashed
       {\npair
          {vp}
          {\semi{0}}}
       {[$\npair
            {v_{i}}
            {\semi{1}}$]}}
  \binatree 
    {cp_{s}} 
    {\unatree 
      {spec} 
      {vp_{j}} 
    } 
    {\slashed{cp_{f}}{[$vp_{j}$]}}
\end{equation}
where $\slashed{vp}{[$v_{i}$]}$ means that the derivation of the
$vp$-node has to include an empty $v$-leaf. In the example in
\figref{vtwo}, the syntactic head (the $c$-position) of the
$cp_{f}$ will be visited before the $vp$ is to be derived, hence the
exact information of the verb trace will be available in time.
Similarly for the movement to the `vorfeld'.
However, if verb second configurations are described by a single
structure
\begin{equation}
\hspace*{-1em}
  \binatree
    {\npair{cp_{s}}{\semi{0}}}
    {\unatree
       {spec}
       {\npair{XP_{i}}{\semi{1}}}
    }
    {\binatree
       {\npair{cp_{f}}{\semi{0}}}
       {\unatree
          {c}
          {v_{j}}
       }
       {\slashed{vp}{[$
          v_{j}$,
          $\npair{XP_{i}}{\semi{1}}$]}}
    }
\end{equation}
the algorithm runs into a deadlock: the $vp$-node cannot be processed
completely, because the semantics of the $XP$-trace is unknown, and
the expansion of the $XP$-filler position will be delayed for the same
reason. If this syntactic description had to be preferred over the one
in (\ref{verbf}), the link relation should be further modified. The
substructure test wrt.\ the semantics of the current goal should be
replaced by a substructure test wrt.\ the global input semantics,
which leads to a loss of flexibility, as it has been discussed in
connection with the pure bottom-up approach.

\subsection{Implementation}
Since the algorithm has been implemented in the CUF language\footnote{The
  CUF-system is an implementation of a theorem prover for a Horn
  clause logic with typed feature terms
  \cite{Doerre/Dorna:1993}.}, which
includes a \pcode{wait}-mechanism, the reodering of subgoals can be
delegated to CUF.

Instead of a full-blown substructure test which might be quite
complicated on graphs like UDRS's, only the predicate names (and other
essential 'semantic' keywords) of the lexical entry are mapped on the
current goal semantics. If such a map is not feasible, this lexical
entry is dropped. 

We restrict the grammars to lexicalized ones. A grammar is lexicalized
if for every local syntax tree there is at least one preterminal leaf,
cf.\ \cite{Schabes/Waters:1993}. Note that lexicalization does not
affect the expressibility of the grammar \cite{Bar-Hillel:1960},
\cite{Schabes/Waters:1993}. However, the generation algorithm turns
much simpler and hence more efficient.  There is no need for a
transitive link relation, since a goal can match immediately the
mother node of a preterminal. The lexicon access and the head-corner
completion step can be merged into one rule schema\footnote{An
  instance of our head-corner generator (without an integrated
  treatment of movement) is the UCG-generator by Calder et al.\ 
  \cite{Calder/Reape/Zeevat:1989} (modulo the use of unary category
  transformation rules) which relies, in addition, on the symmetry of
  syntactic and semantic head.  A syntactic-head-driven generator for
  a kind of lexicalized grammars has been proposed
  independently by 
  \cite{Kay:1993mts}.  Another
  variant of a lexicalized grammar by 
  \cite{Dymetman/Isabelle/Perrault:1990} does not make use of the
  head-corner idea but rather corresponds to the top-down generation
  schema presented in \figref{tdgen}.}. 

\begin{figure*}
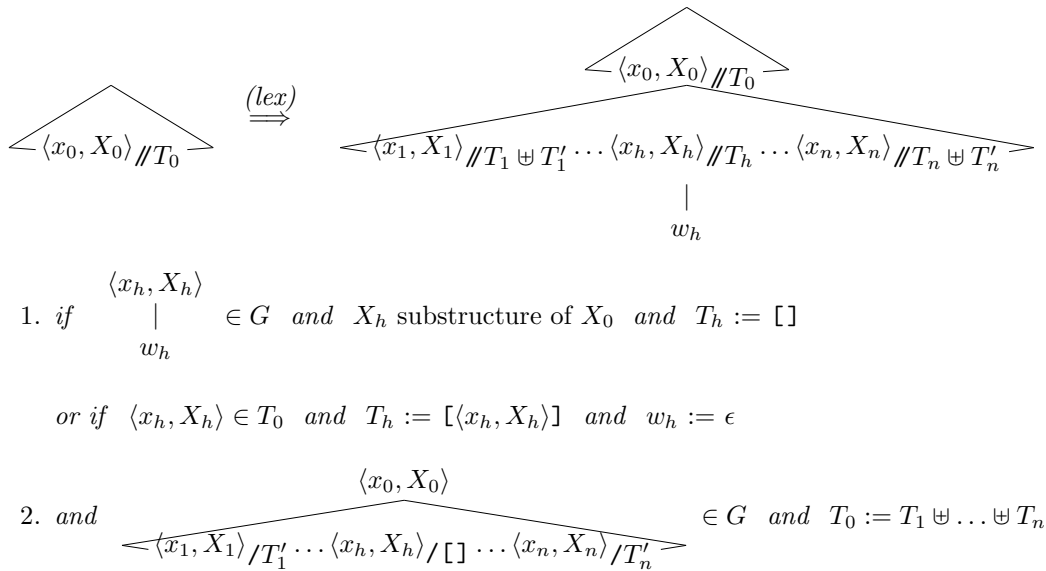

all leaves are labeled with terminals
\hspace{2em}
\rname{success} 
\\
\begin{math}
\infrule
  {
   \vddtree
     {\dslashed{\npair{\categi{0}}{\semi{0}}}{T_{0}}}
  }
  {
    \begin{array}{c}
        \vddtree
          {\dslashed{\npair{\categi{0}}{\semi{0}}}{T_{0}}} \\[-0.8em]
        \vddtree
          {\dslashed{\npair{\categi{1}}{\semi{1}}}{T_{1}\uplus T_{1}'}
           \ldots
           \dslashed{\npair{\categi{h}}{\semi{h}}}{T_{h}}
           \ldots
           \dslashed{\npair{\categi{n}}{\semi{n}}}{T_{n}\uplus T_{n}'}} \\[-0.8em]
        \unatree
          {}
          {\lexemi{h}}
      \end{array}
   }
   {lex} 
\end{math}
\begin{enumerate}
  \item
    {\em if\/\ }
        $\unatree
           {\npair{\categi{h}}{\semi{h}}}
           {\lexemi{h}}
        \in \grami{}$ {\em \ and\/\ }
            $\semi{h}$ substructure of $\semi{0}$ {\em \ and\/\ }
            $T_{h}$ := \pcode{[\hspace{0.1em}]}

       \vspace*{1em}
       {\em or if\/\ }
       $\npair{\categi{h}}{\semi{h}} \in T_{0}$ {\em \ and\/\ }
           $T_{h}$ := \pcode{[$\npair{\categi{h}}{\semi{h}}$]} {\em \ and\/\ } 
           $\lexemi{h}$ := $\epsilon$
       \\[-0.5em]
  \item
   {\em and \/}
    $  \vdtree
         {\npair{\categi{0}}{\semi{0}}}
         {\slashed{\npair{\categi{1}}{\semi{1}}}{$T_{1}'$}
          \ldots
          \slashed{\npair{\categi{h}}{\semi{h}}}{\pcode{[\hspace{0.1em}]}}
          \ldots
          \slashed{\npair{\categi{n}}{\semi{n}}}{$T_{n}'$}
         }
     \in \grami{}$ {\em \ and\/\ }
     $T_{0} := T_{1} \uplus \ldots \uplus T_{n}$
\end{enumerate}
\caption{Head-Corner Generator for lexicalized grammars
  ($\grami{}$ grammar description; 
   $\categi{i}$ syntactic category symbol;
   $\semi{i}$ semantic representation;
   $T_{i}$ {\tt slash}-values)}
\label{lhcgen}
\end{figure*}

A version of the Non-Local-Feature principle of HPSG 
has been integrated into the algorithm. Every non-head nonterminal
leaf of a local tree must come with a (possibly empty) multiset of
syntax-semantics pairs as the value of its
\pcode{to\_bind:slash}-feature (feature abbreviated as \pcode{/}),
cf.\ example (\ref{verbf}).  From these static values, the dynamic
\pcode{inherited:slash}-values (feature abbreviated as
\pcode{/$\!\!$/}) can be calculated during generation, see
rule~\rname{lex} in \figref{lhcgen}.

    {\bf (1a)} Choose a lexical entry as the head~$\categi{h}$ of the
        current goal~$\categi{0}$. Then the substructure condition must
        hold for the corresponding semantic forms $\semi{h}$ and $\semi{0}$.
        The \pcode{/$\!\!$/}-value~$T_{h}$ must be empty.

     {\bf (1b)}  Or choose an element of the \pcode{/$\!\!$/}-value~$T_{0}$
        of the current head~$\categi{0}$. 
        Then the \pcode{/$\!\!$/}-value~$T_{h}$ becomes
        \pcode{[$\npair{\categi{h}}{\semi{h}}$]}. The associated
        string~$\lexemi{h}$ is empty. 

 {\bf (2)}  There must be a lexicalized tree which connects the
    goal~$\categi{0}$ and the chosen head~$\categi{h}$.
    The \pcode{/$\!\!$/}-value~$T_{0}$ is split 
    into disjoint sets $T_{1}$, \ldots, $T_{n}$.
    The \pcode{/$\!\!$/}-values of the new
    subgoals~$\categi{1}$, \ldots, $\categi{n}$ are the
    disjoint set unions $T_{i} \uplus T_{i}'$ where $T_{i}'$ is the
    \pcode{/}-value of $\categi{i}$ in the local tree given in the grammar.

Note that this version of the Non-Local-Feature principle corresponds to
the hypothetical reasoning mechanism which is provided by the Lambek
categorial grammars \cite{Lambek:1958}, \cite{Koenig:1994hrala}. This
is illustrated by the fact that e.g.\ the left tree in
example (\ref{verbf}) can be rendered in categorial grammar notation as
$cp_{f}/(vp/v)$. Hence, the algorithm in \figref{lhcgen} has a clear
logical basis.


\section{Conclusion}
This paper gives a syntactic-head-driven generation algorithm which
includes a well-defined treatment of moved constituents. Since it
relies on the notion of a syntactic head instead of a semantic head it
works also for grammars where semantic heads are not available in
general, like for a grammar which includes semantic decomposition
rules of (scopally) Underspecified Discourse Representation
Structures.  By using the same notion of head both for parsing and for
generation, both techniques become even closer. In effect, the
abstract specifications of the generation algorithms which we gave
above, could be read as parsing algorithms, modulo a few changes (of
the success condition and the link relation).

Generation from Underspecified DRS's means that sentences can be
generated from meaning representations which have not been
disambiguated with regard to quantifier scope. This is of particular
importance for applications in machine translation, where one wants
to avoid the resolution of scope relations as long as the
underspecified meaning can be rendered in the source and in the target
language. Future work should consider more the strategic part of the
generation problem, e.g.\ try to find heuristics and strategies which
handle situations of `scope mismatch' where one language has to be
more precise with regard to scope than the other.




\begin{footnotesize}

\end{footnotesize}

\end{document}